\documentclass{article}
\usepackage{spconf,amsmath,epsfig,amssymb}
\usepackage{multirow}
\usepackage{booktabs}
\usepackage{graphicx, color}
\usepackage[normalem]{ulem}

\let\OLDthebibliography\thebibliography
\renewcommand\thebibliography[1]{
  \OLDthebibliography{#1}
  \setlength{\parskip}{0pt}
  \setlength{\itemsep}{0pt plus 0.3ex}
}

\newcommand\hui[1]{\textcolor{black}{#1}}

\newcommand\XP[1]{\textcolor{black}{#1}}

\begin{document}\sloppy

\title{Object Counting: You Only Need to Look at One}
%
\name{Hui LIN, Xiaopeng HONG, Yabin WANG}
\address{\small School of Cyber Science and Engineering, Xi'an Jiaotong University, China\\
\small Emails: {linhuixjtu@gmail.com;  hongxiaopeng@ieee.org; iamwangyabin@stu.xjtu.edu.cn}}


\maketitle

\begin{abstract}

This paper aims to tackle the challenging task of one-shot object counting. Given an image containing novel, previously unseen category objects, the goal of the task is to count all instances in the desired category with only one supporting bounding box example. To this end, we propose a counting model by which you only need to Look At One instance (LaoNet). First, a feature correlation module combines the Self-Attention and Correlative-Attention modules to learn both inner-relations and inter-relations. It enables the network to be robust to the inconsistency of rotations and sizes among different instances. Second, a Scale Aggregation mechanism is designed to help extract features with different scale information. Compared with existing few-shot counting methods, LaoNet achieves state-of-the-art results while learning with a high convergence speed. The code will be available soon.


\end{abstract}

\begin{figure*}[t]
    \centering
    \includegraphics[width = 0.95\textwidth]{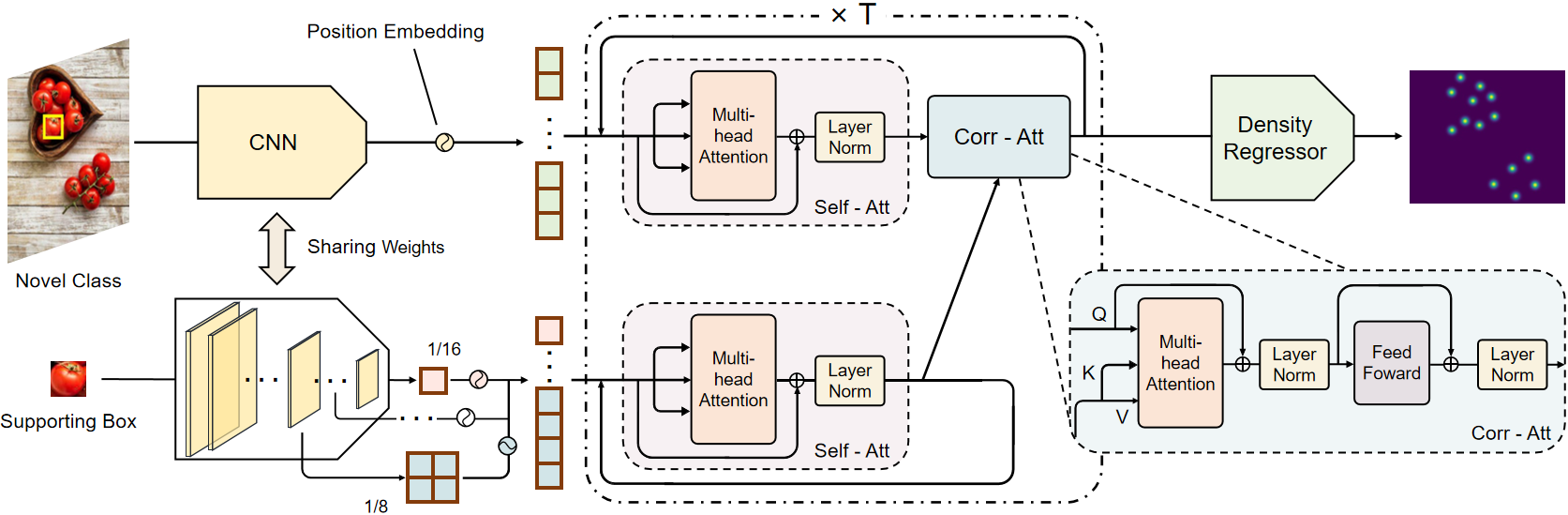}
    \caption{The overall architecture of the proposed LaoNet for one-shot object counting. Both the query image and the supporting box are fed into CNN to extract features. Supporting features are aggregated among scales. Then the flatten features with unique position embedding are transmitted into feature correlation model with Self-Attentions and Correlative Attentions. Finally, a density regressor is adopt\XP{ed} to predict the final density map. }
    \label{fig:network}
\end{figure*}


\section{Introduction}

Object counting has become increasingly important due to its wide range of applications such as crowd surveillance, traffic monitoring, wildlife conservation and inventory management. Most of the existing counting methods~\cite{zhang2016single, ma2019bayesian, biswas2017automatic} focus on a \XP{particular,} single category. However, when applying them into new categories, their performances will drop catastrophically. Meanwhile, it is \XP{extremely} difficult and costly to collect all categories and label them for training. 

For humans, the generalization ability allows them to learn and deal with various vision tasks without much prior knowledge and experience. We are amazed by this remarkable ability and in this work, we focus on this learning paradigm and design a network to efficiently recognize and count new categories given only one example. We follow the few-shot setting in ~\cite{ranjan2021learning} and modify it to one-shot object counting. That is, the model takes an image with unseen novel categories and a supporting bounding box containing an example instance of desired category as input, and then predicts the object count in the image. 

However, there are two main challenges. \textbf{First}, the object counting task includes many different categories, and even several categories exist within a same image. Moreover in few-shot setting, these categories will not overlap between training and inference. This means that the model needs to have a strong distinguishing ability between features of different categories, and meanwhile, an effective associating ability among instances sharing the same category.
\textbf{Second}, in one-shot counting, the model \XP{learns}
from only one supporting instance. 
Much of the difficulty results from the fact that the supporting sample may \XP{differ from other instances in, for example, sizes and poses. Hence, the model is required to be invariant towards these variations without seeing the commonalities across different instances.}

Therefore, in this paper, we propose an effective network named LaoNet for one-shot object counting. It consists of three main parts: feature extraction, feature correlation and the density regressor, as shown in Figure~\ref{fig:network}. The feature correlation model and the feature extraction model are elaborately designed to address the above two challenges.

We propose the feature correlation based on Self-Attention and Correlative-Attention modules to learn inner-relations and inter-relations respectively. The Self-Attention encourages the model to focus more on important features and their correlations, improving the efficiency of information refinement. Previous few-shot counting methods~\cite{ranjan2021learning, yang2021class} usually leverage on a convolution operation to match the similarities between image features and supporting features. However, as the kernel is derived from supporting features with the default size and rotation angle, the convolution operation will greatly depend on the quality of supporting features and the consistency of physical properties among different instances. Instead, our designed feature correlation model benefits from two kinds of attention modules and addresses the above problem by considering all correlations.

We further propose a Scale Aggregation mechanism in scale extraction to deal with scale variations among different categories and different instances. By learning features from multi-subspace, the model aggregates various scale information while maintaining a spatial consistency.

To summarize, our contribution is threefold. 

\begin{itemize}
    \item We design a novel network named LaoNet (A network by which you only need to Look At One instance) for one-shot object counting. By combining Self-Attention and Correlative-Attention modules,  LaoNet exploits the correlation among novel category objects with high accuracy and efficiency.
    
	\item We propose a Scale Aggregation mechanism to extract more comprehensive features and fuse multi-scale information from the supporting box.
	
	\item The experimental results show that our model achieves state-of-the-art results with significant improvements on FSC-147~\cite{ranjan2021learning} and COCO~\cite{lin2014microsoft} datasets under the one-shot setting without fine-tuning.
	
	
\end{itemize}


\section{Related Works}

Object counting methods can be briefly divided into two types. Detection based methods~\cite{chattopadhyay2017counting} count the number of objects by exhaustively detecting every target in images. But they rely on the complex labels such as bounding boxes. Regression based methods~\cite{zhang2016single, ma2019bayesian} learn to count by predicting a density map, in which each value represents the density of target objects at the corresponding location. The count prediction equals to the total sum of density map.

Nevertheless, most of the counting methods are category specifically, e.g. for human crowd~\cite{zhang2016single, ma2019bayesian, cao2018scale, liu2019context, wang2020distribution, lin2021direct}, for cars~\cite{biswas2017automatic, moranduzzo2013automatic}, for plants~\cite{machefer2020mask} or for cells~\cite{falk2019u, xie2018microscopy}. They focus on only one category and will loss the original satisfied performance when transferring to other categories. \XP{Moreover, most traditional approaches usually rely on tens of thousands of instances to train a counting model~\cite{ma2019bayesian, cao2018scale, liu2019context, lin2021direct, biswas2017automatic, moranduzzo2013automatic}.} 

\XP{To reduce considerably the number of samples needed to train a counting model} \hui{for a particular category}, recently, few-shot counting task has been developed. The key lies in the generalization ability of the model to deal with novel categories from few labeled examples. The study~\cite{lu2018class} proposes a Generic Matching Network (GMN) for class-agnostic counting. However it still needs several dozens to hundreds examples of a novel category for adaptation and good performance. CFOCNet is introduced to match and utilize the similarity between objects within the same category~\cite{yang2021class}. The work~\cite{ranjan2021learning} presents a Few Shot Adaptation and Matching Network (FamNet) to learn feature correlations and few-shot adaptation \XP{and} also introduces a few-shot counting dataset named FSC-147.

When the number of labeled example decreases to one, the task evolves into one-shot counting. In other visual tasks, researchers develop methods for one-shot segmentation~\cite{michaelis2018one} and one-shot object detection~\cite{hsieh2019one, li2020one}. Compared to the few-shot setting \XP{which usually uses at least three instances for each object~\cite{ranjan2021learning}, the one-shot setting, where only one instance is available, is clearly more challenging.} 

\XP{It is worth mentioning that detection based approaches \cite{lin2017focal, ren2015faster, he2017mask} are inferior for the tasks of few-shot and one-shot counting. One main reason is that it requires extra and costly bounding-box annotations of \emph{all} instances in the training stage while \XP{one-shot counting approach which we focus on} depends on dot annotations and only \emph{one} supporting box. To illustrate this point further, 
we perform experiments in Section~\ref{sec:comparison} to compare with detection based approaches and validate the proposed network for one-shot counting.}


\section{Approach}

\subsection{Problem Definition}

One-shot object counting consists of a training set $(I_t, s_t, y_t) \in \mathcal{T}$ and a query set $(I_q, s_q) \in \mathcal{Q}$, in which categories are mutually exclusive. Each input for the model contains an image $I$ and a supporting bounding box $s$ annotating one object of the desired category. In training set, abundant point annotations $y_t$ are available to supervise the model. In inference stage, we aim the model to learn to count the novel objects in $I_q$ with a supporting category instance sampled by $s_q$.

\subsection{Feature Correlation}
As the model is required to learn to count from only one supporting object, seizing the correlation between features with high efficiency is quite important. Therefore, we build the feature correlation model in our one-shot network based on Self-Attention and Correlative-Attention modules, for learning the inner-relations and inter-relations respectively. 

As illustrated in Figure~\ref{fig:network} (violet block), our Self-Attention module consists of a Multi-head Attention (MA) and a layer normalization (LN). We first introduce the definition of attention~\cite{vaswani2017attention}, given the query $Q$, key $K$ and value vector $V$:
\begin{equation}
    A(Q, K, V\ |\ W) = \mathcal{S}(\frac{(QW^Q)(KW^K)^T}{\sqrt{d}}+PE)(VW^V),
\end{equation}
where $\mathcal{S}$ is the softmax function and $\frac{1}{\sqrt{d}}$ is a scaling factor based on the vector dimension $d$. $W:W^Q, W^K, W^V \in \mathbb{R}^{d \times d}$ are weight matrices for projections and $PE$ is the position embedding.

To leverage on more representation subspaces, we adopt the extending form with multi attention heads:
\begin{equation}
\begin{aligned}
    MA(Q, K, V) = Concat(head_1,.., head_h)W^O \\
    where\ head_i = A(Q, K, V\ |\ W_i).
\end{aligned}
\end{equation}
The representation dimensions are divided by parallel attention heads, where parameter matrices $W_i:W_i^Q, W_i^K, W_i^V \in \mathbb{R}^{d \times d/h}$ and $W_O \in \mathbb{R}^{d \times d}$.

One challenging problem in counting task is the existence of many complex interfering things. To efficiently weaken the negative influence by those irrelevant background, we apply Multi-head Self-Attention in image features to learn inner-relations and encourage the model to focus more on repetitive objects that can be counted.

We denote the feature sequences of the query image and the supporting box region as $X$ and $S$, with sizes $X \in \mathbb{R}^{HW \times C}$ and $S \in \mathbb{R}^{hw \times C}$. And the refined query feature is calculated by:
\begin{equation}
    \tilde{X} = LN(MA(X_Q,X_K,X_V)+X).
\end{equation}
A layer normalization (LN) is adopted to balance the value scales.

Meanwhile, as there is only one supporting object in one-shot counting problem, refining the salient features within the object is necessary and helpful for counting efficiency and accuracy. Therefore we apply another Self-Attention module to supporting feature and get refined $\tilde{S}$.

Previous few-shot counting methods~\cite{ranjan2021learning, yang2021class} usually adopt a convolution operation where the supporting features act as kernels to match the similarities for target category. However, the results will greatly depend on the quality of supporting features and the consistency of objects' properties, including rotations and scales.

To this end, we propose a Correlative-Attention module to learn inter-relations between query and supporting features and alleviate the constraints of irrelevant properties.

Specifically, we extend the MA by learning correlations between different feature sequences and add a feed-forward network (FFN) to fuse the features, i.e.,
\begin{equation}
    X^* = Corr(\tilde{X}, \tilde{S}) = \mathcal{G} (MA(\tilde{X}_Q,\tilde{S}_K,\tilde{S}_V)+\tilde{X}).
\end{equation}
$\mathcal{G}$ includes two LNs and a FFN in the form of residual (light blue block in Figure~\ref{fig:network}). Finally, $X^*$ and $\tilde{S}$ will be fed into the cycle as new feature sequences where each cycle consists of two Self-Attention modules and a Correlative-Attention module.

\subsection{Feature Extraction and Scale Aggregation}

To extract feature sequences from images, we use VGG-19 as our backbone. For query image, the output of the final level is directly flattened and transmitted into Self-Attention module. For the supporting box, as there are uncontrollable scale variations among instances due to the perspective, we propose a Scale Aggregation mechanism to fuse different scale information.

Given $l$ as the number of layers in CNN, we aggregate the feature maps among different scales:
\begin{equation}
    S = Concat(\mathcal{F}^l(s), \mathcal{F}^{l-1}(s),...,\mathcal{F}^{l+1-\delta}(s)),
\end{equation}
where $\mathcal{F}^i$ represents a feature map at $i_{th}$ level and $\delta \in [1, l]$ decides the number of layers taken for aggregation.

Meanwhile, we leverage on identifying position embedding to help the model distinguish the integrated scale information in attention model. By adopting the fixed sinusoidal absolute position embedding~\cite{vaswani2017attention}, feature sequences from different scales can still maintain the consistency between positions, i.e., 
\begin{equation}
\begin{aligned}
    &PE_{(pos_j, 2i)} = sin(pos_j / 10000^{2i/d}),\\
    &PE_{(pos_j, 2i+1)} = cos(pos_j / 10000^{2i/d}).
\end{aligned}
\end{equation}
$i$ is the dimension and $pos_j$ is the position for $j_{th}$ feature map.

\subsection{Training Loss}

We use Euclidean distance to measure the difference between estimated density map and ground truth density map, which is generated based on annotated points following ~\cite{zhang2016single}. The loss is defined as follows:
\begin{equation}
    \mathcal{L}_E = ||D^{gt}-D||^2_2,
\end{equation}
where $D$ is the estimated density map and $D^{gt}$ is the ground truth density map. To improve the local pattern consistency, we also adopt a SSIM loss followed the calculation in ~\cite{cao2018scale}. By integrating the above two loss functions, we have
\begin{equation}
    \mathcal{L} = \mathcal{L}_E + \lambda \mathcal{L}_{SSIM},
\end{equation}
where $\lambda$ is the balanced weight.

\section{Experiments}

\subsection{Implement Details and Evaluation Metrics}
We design the density regressor by an upsampling layer and three convolution layers with ReLU activation. The kernel sizes of first two layers are 3 × 3 and that of last is 1 × 1. Random scaling and flipping are adopt\XP{ed} for each training image. Adam~\cite{kingmaadam} with a learning rate $0.5 \times 10^{-5}$ is used to optimize the parameters. We set the number of attention heads $h$ as 4, the correlation cycle $T$ as 2, the number of aggregated layers $\delta$ as 2, and the loss balanced parameter $\lambda$ as $10^{-4}$.

Mean Absolute Error (MAE) and Root Mean Squared Error (RMSE) are used to measure the performance of our methods. They are defined by: 
\begin{equation}
\begin{aligned}
    &MAE=\frac{1}{M}\ \sum_{i=1}^{M}\big|N_i^{gt}-N_i\big|,\\ &RMSE=\sqrt{\frac{1}{M}\ \sum_{i=1}^{M}(N_i^{gt}-N_i)^2)},
\end{aligned}
\end{equation}
where \XP{$M$ and $N^{gt}$ are the number of images and the ground-truth count, respectively}. The predicted count $N$ is calculated by \XP{integrating the} estimated density map $D$.

\subsection{Datesets}
\noindent \textbf{FSC-147~\cite{ranjan2021learning}} contains a total of 6135 images collected for few-shot counting problem. In each image, three randomly selected object instances are annotated by bounding boxes while other instances are annotated by points. 89 object categories with 3,659 images are divided for training set. Each 29 categories with 1,286 and 1,190 images respectively are divided for validation and testing sets.

\noindent \textbf{MS-COCO~\cite{lin2014microsoft}} is a large dataset widely used in object detection and instance segmentation. In val2017 set, there are 80 common object categories with 5,000 images in complex everyday scenes. We follow~\cite{michaelis2018one} to generate four train/test splits which each contains 60 training and 20 testing categories.

\renewcommand{\tabcolsep}{8 pt}{
\begin{table}[t]
\small
\begin{center}
\begin{tabular}{lcccc}
  \toprule[1pt]
  \multirow{2}*{Methods} & \multicolumn{2}{c}{Val} & \multicolumn{2}{c}{Test}\\
  & MAE & RMSE & MAE & RMSE \\
  \hline
  \textit{3-shot} & & & & \\
  Mean & 53.38 & 124.53 & 47.55 & 147.67 \\
  Median & 48.68 & 129.70 & 47.73 & 152.46 \\
  FR detector~\cite{kang2019few} & 45.45 & 112.53 & 41.64 & 141.04 \\
  FSOD detector~\cite{fan2020few} & 36.36 & 115.00 & 32.53 & 140.65 \\
  GMN~\cite{lu2018class} & 29.66 & 89.81 & 26.52 & 124.57 \\
  MAML~\cite{finn2017model} & 25.54 & 79.44 & 24.90 & 112.68 \\
  FamNet~\cite{ranjan2021learning} & 23.75 & 69.07 & 22.08 & 99.54 \\
  \hline
 \textit{1-shot} & & & & \\
  CFOCNet~\cite{yang2021class} & 27.82 & 71.99 & 28.60 & 123.96 \\
  FamNet~\cite{ranjan2021learning} & 26.55 & 77.01 & 26.76 & 110.95 \\
  LaoNet (Ours) & \textbf{17.11} & \textbf{56.81} & \textbf{15.78} & \textbf{97.15} \\
  \toprule[1pt]
\end{tabular}
\caption{Comparisons with previous state-of-the-art few-shot methods on FSC-147. The upper part of the table presents the results in 3-shot setting while the lower part presents 1-shot results. FamNet~\cite{ranjan2021learning} uses the adaptation strategy during testing. It is worth noticing that our one-shot LaoNet outperforms all of previous methods, even those in 3-shot setting, without any fine-tuning strategy.} \label{tab:fam}
\end{center}
\end{table}}

\renewcommand{\tabcolsep}{4 pt}{
\begin{figure}[t!]
	\begin{center}
		\begin{tabular}{ccc}
			\includegraphics[height=0.22\linewidth]{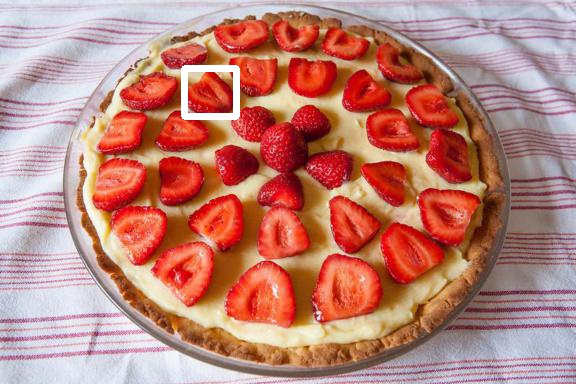}  &
			\includegraphics[height=0.22\linewidth]{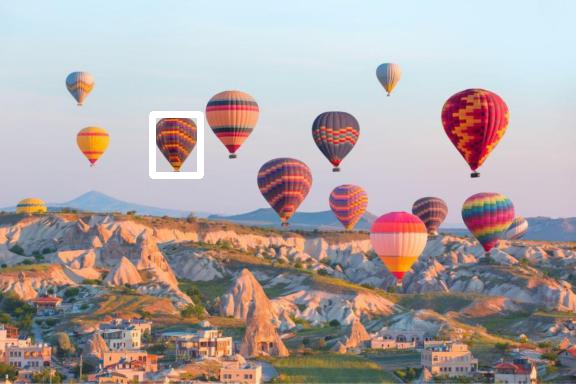}  &
			\includegraphics[height=0.22\linewidth]{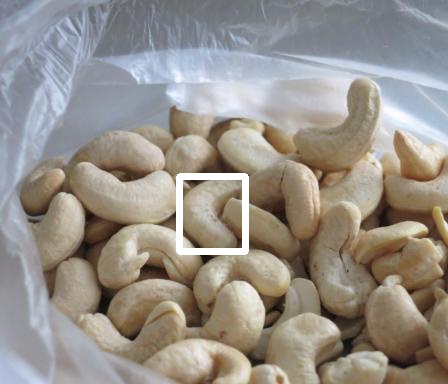}  \\
			\footnotesize{GT: 33} & \footnotesize{GT: 14} & \footnotesize{GT: 35}  \\
			\includegraphics[height=0.22\linewidth]{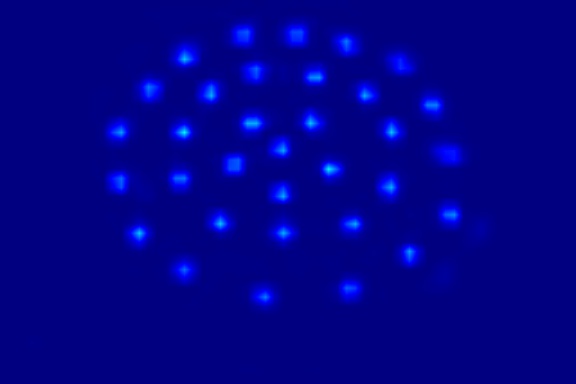}  &
			
			\includegraphics[height=0.22\linewidth]{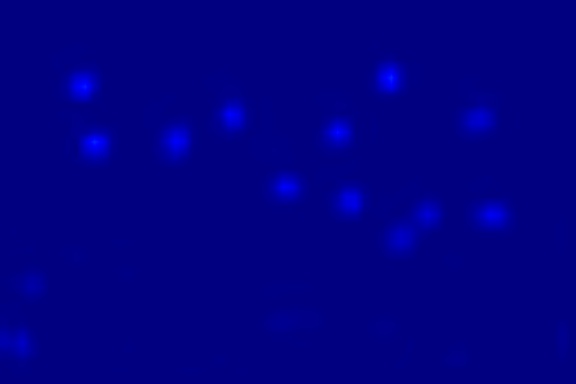}  &
			\includegraphics[height=0.22\linewidth]{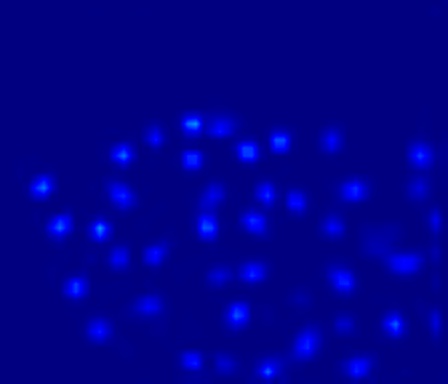} \\
			\footnotesize{Pre: 35} & \footnotesize{Pre: 14} & \footnotesize{Pre: 37}  \\

		\end{tabular}
		\caption{Visualizations of one-shot counting inputs and corresponding predicted density maps. The model can perform great counting accuracy even it has never seen strawberry, hot air balloon or cashew before.}
		\label{fig:vis}
	\end{center}
\end{figure}}

\subsection{Comparison with Few-Shot Approaches} \label{sec:comparison}

\renewcommand{\tabcolsep}{10 pt}{
\begin{table*}[t]
\small
\begin{center}
\begin{tabular}{lcccccccccc}
  \toprule[1pt]
  \multirow{2}*{Methods} & \multicolumn{2}{c}{Fold\ 0} & \multicolumn{2}{c}{Fold\ 1} & \multicolumn{2}{c}{Fold\ 2} & \multicolumn{2}{c}{Fold\ 3} & \multicolumn{2}{c}{Average}\\
  & MAE & RMSE & MAE & RMSE & MAE & RMSE & MAE & RMSE & MAE & RMSE\\
  \hline
  Segment~\cite{michaelis2018one}$^\dag$ & 2.91 & 4.20 & 2.47 & 3.67 & 2.64 & 3.79 & 2.82 & 4.09 & 2.71 & 3.94 \\
  GMN~\cite{lu2018class}$^\dag$ & 2.97 & 4.02 & 3.39 & 4.56 & 3.00 & 3.94 & 3.30 & 4.40 & 3.17 & 4.23 \\
  CFOCNet~\cite{yang2021class}$^\dag$ & 2.24 & \textbf{3.50} & 1.78 & 2.90 & 2.66 & 3.82 & 2.16 & 3.27 & 2.21 & 3.37 \\
  \hline
  FamNet~\cite{ranjan2021learning} & 2.34 & 3.78 & 1.41 & 2.85 & 2.40 & 2.75 & 2.27 & 3.66 & 2.11 & 3.26 \\
  CFOCNet~\cite{yang2021class} & 2.23 & 4.04 & 1.62 & 2.72 & 1.83 & 3.02 & 2.13 & 3.03 & 1.95 & 3.20 \\
  LaoNet (Ours) & \textbf{2.20} & 3.78 & \textbf{1.32} & \textbf{2.66} & \textbf{1.58} & \textbf{2.19} & \textbf{1.84} & \textbf{2.90} & \textbf{1.73} & \textbf{2.93} \\
  \toprule[1pt]
\end{tabular}
\caption{Results on each of four folds of COCO val2017. Methods with $\dag$ follow the experiment setting in ~\cite{yang2021class}. Our method achieves great accuracy without any fine-tuning on testing categories.} \label{tab:coco}
\end{center}
\end{table*}}

We hold experiments on above two few-shot counting datasets to evaluate the proposed network. \XP{As there are few existing methods specifically designed for one-shot counting, for comprehensive evaluation, we modify FamNet~\cite{ranjan2021learning} and CFOCNet~\cite{yang2021class} for this setting and also compare with other few-shot counting approaches~\cite{kang2019few, fan2020few, lu2018class, finn2017model, michaelis2018one}.}

First, quantitative results on FSC-147 are shown in Table~\ref{tab:fam}. We list seven results of previous few-shot detection and counting methods in 3-shot setting and two results of state-of-the-art counting methods in 1-shot setting for comparison. The result of FamNet~\cite{ranjan2021learning} uses the adaptation strategy during testing.

It is worth noticing that our one-shot LaoNet outperforms all of previous few-shot methods, even those in 3 shot setting, without any fine-tuning strategy. We have generated new records by reducing the error of FamNet from 26.55 to 17.11 for MAE and from 77.01 to 56.81 for RMSE in validation set, from 26.76 to 15.78 for MAE and from 110.95 to 97.15 for RMSE in testing set.

Second, Table~\ref{tab:coco} shows the results on each of four folds of COCO val2017. Methods with $\dag$ in the upper part of the table follow the experiment setting in ~\cite{yang2021class}. That is, the supporting examples are chosen from all instances in the dataset during training and testing, which is laborious and costly under the need of all instances annotated by bounding boxes. While our setting allows only one fixed instance for each image, we re-conduct the experiment of CFOCNet~\cite{yang2021class}. As the result shows, our method maintains a great performance on COCO dataset.

\subsection{Discussions}

\renewcommand{\tabcolsep}{6 pt}{
\begin{table}[t]
\small
\begin{center}
\begin{tabular}{lcccc}
  \toprule[1pt]
  \multirow{2}*{Methods} & \multicolumn{2}{c}{Val} & \multicolumn{2}{c}{Test} \\
   & MAE & RMSE & MAE & RMSE \\
  \hline
  LaoNet & \textbf{17.11} & \textbf{56.81} & \textbf{15.78} & \textbf{97.15} \\
  $-$ Self-Attention ($X$) & 19.83 & 64.84 & 19.71 & 107.32 \\
  $-$ Self-Attention ($S$) & 19.67 & 63.79 & 18.71 & 111.83 \\
  $-$ Scale Aggregation & 18.82 & 63.74 & 17.16 & 106.40 \\
  $-$ SSIM & 17.82 & 57.66 & 16.11 & 100.59 \\
  \toprule[1pt]
\end{tabular}
\caption{Ablation study for different terms. $X$ stands for feature sequences of query image and $S$ stands for that of supporting box region. Experiments are performed in FSC-147 val and test.} \label{tab:ablation}
\end{center}
\end{table}}

\renewcommand{\tabcolsep}{5 pt}{
\begin{table}[t]
\small
\begin{center}
\begin{tabular}{lcccc}
  \toprule[1pt]
  \multirow{2}*{Methods} & \multicolumn{2}{c}{FSC147-COCO Val} & \multicolumn{2}{c}{FSC147-COCO Test}\\
  & MAE & RMSE & MAE & RMSE \\
  \hline
  RetinaNet~\cite{lin2017focal} & 63.57 & 174.36 & 52.67 & 85.86 \\
  Faster R-CNN~\cite{ren2015faster} & 52.79 & 172.46 & 36.20 & 79.59 \\
  Mask R-CNN~\cite{he2017mask} & 52.51 & 172.21 & 35.56 & 80.00 \\
  \hline
  FamNet~\cite{ranjan2021learning} & 39.82 & 108.13 & 22.76 & 45.92 \\
  LaoNet (Ours) & \textbf{31.12} & \textbf{97.15} & \textbf{12.89} & \textbf{26.64} \\
  \toprule[1pt]
\end{tabular}
\caption{Comparisons with pre-trained object detectors on FSC147-COCO splits of FSC147 which contain images with COCO categories. Even pre-trained with thousands of annotated examples on MS-COCO dataset, these object detectors still perform unsatisfied accuracy on counting task.} \label{tab:detect}
\end{center}
\end{table}}

\noindent \textbf{Contribution of Different Terms.} We study the accuracy contributions of different terms in FSC-147. The result is shown in Table~\ref{tab:ablation}, \XP{each row whereof reports the results after removing one component or one term from LaoNet.}
The Self-Attention modules for the two feature sequences to learn inner-relations increase the accuracy in testing set by $19.9\%$ and $15.7\%$ for MAE, $9.5\%$ and $13.1\%$ for RMSE, respectively. Compared to other two terms, the Self-Attention modules contribute most to the performance of our model.

The Scale Aggregation mechanism helps more on RMSE. The result demonstrates a robustness contribution under the multi-scale aggregation. Finally, the SSIM loss further improves the counting accuracy by both lower MAE and RMSE.

\noindent \textbf{Convergence Speed.} We hold experiments to measure the convergence speed and the performance stability. We pick FamNet~\cite{ranjan2021learning} as the baseline for LaoNet \XP{with a pre-trained CNN backbone and an Adam optimizer.} We train both two models on FSC-147 and report the validation MAE \XP{for} 100 epochs. 

\begin{figure}[t]
  \centering
  \includegraphics[width = 0.48\textwidth]{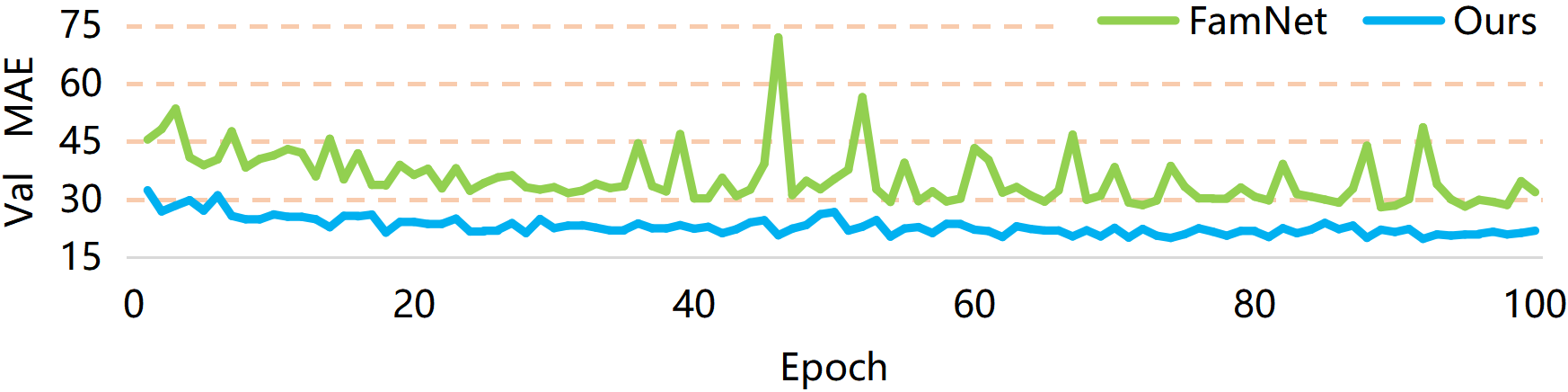}
   \caption{Comparisons of validation MAE during training. The blue line represents our proposed LaoNet. With just one epoch, it can perform a great accuracy which FamNet needs to train for about 20 epochs.}
   \label{fig:converge}
\end{figure}

As shown in Figure~\ref{fig:converge}, our model \XP{has} faster convergence speed and \XP{better} stability \XP{than FamNet}. With just \XP{2 epoches}, our method achieves a low counting error which FamNet \XP{has to reach after 40} epochs. Meanwhile, \XP{the convergence of our method is smooth and stable, while that of Famet is jagged, with multiple sharp peaks and the highest error of 70.}


\noindent \textbf{Comparison with Object Detectors.}
Object detectors can be used for counting task with the number of predicted detections. However, even these detectors work with categories which they are trained on instead of one-shot setting, their counting performances are still limited. We select images of FSC-147-COCO subset from FSC147 Val and Test sets which share categories with MS-COCO dataset and conduct quantitative experiments.

As the results shown in Table~\ref{tab:detect}, we compare LaoNet with several object detectors which are well pre-trained with thousands of annotated examples on MS-COCO. Nevertheless, our method, which counts \XP{\emph{unseen}} categories, still outperforms the detection based methods which \emph{have met} those categories in training, by a large margin.

\section{Conclusion}

This paper targets one-shot object counting, which requires the counting model to count objects of new categories by looking at only one instance. We propose an efficient network named LaoNet to address this challenge. LaoNet includes a feature correlation module to learn both inner-relations and inter-relations and a scale aggregation module to extract multi-scale information for improving robustness. Without any fine-tuning in inference, our LaoNet outperforms previous state-of-the-art few-shot counting methods with a high convergence speed. In the future, we consider applying our model to a wider range of one-shot vision tasks.


\bibliographystyle{IEEEbib}
\bibliography{main.bib}

\begin{thebibliography}{10}

\bibitem{zhang2016single}
Yingying Zhang, Desen Zhou, Siqin Chen, Shenghua Gao, and Yi~Ma,
\newblock ``Single-image crowd counting via multi-column convolutional neural
  network,''
\newblock in {\em CVPR}, 2016.

\bibitem{ma2019bayesian}
Zhiheng Ma, Xing Wei, Xiaopeng Hong, and Yihong Gong,
\newblock ``Bayesian loss for crowd count estimation with point supervision,''
\newblock in {\em ICCV}, 2019.

\bibitem{biswas2017automatic}
Debojit Biswas, Hongbo Su, Chengyi Wang, Jason Blankenship, and Aleksandar
  Stevanovic,
\newblock ``An automatic car counting system using overfeat framework,''
\newblock {\em Sensors (Basel)}, 2017.

\bibitem{ranjan2021learning}
Viresh Ranjan, Udbhav Sharma, Thu Nguyen, and Minh Hoai,
\newblock ``Learning to count everything,''
\newblock in {\em CVPR}, 2021.

\bibitem{yang2021class}
Shuo-Diao Yang, Hung-Ting Su, Winston~H Hsu, and Wen-Chin Chen,
\newblock ``Class-agnostic few-shot object counting,''
\newblock in {\em WACV}, 2021.

\bibitem{lin2014microsoft}
Tsung-Yi Lin, Michael Maire, Serge Belongie, James Hays, Pietro Perona, Deva
  Ramanan, Piotr Doll{\'a}r, and C~Lawrence Zitnick,
\newblock ``Microsoft coco: Common objects in context,''
\newblock in {\em ECCV}. Springer, 2014.

\bibitem{chattopadhyay2017counting}
Prithvijit Chattopadhyay, Ramakrishna Vedantam, Ramprasaath~R Selvaraju, Dhruv
  Batra, and Devi Parikh,
\newblock ``Counting everyday objects in everyday scenes,''
\newblock in {\em CVPR}, 2017, pp. 1135--1144.

\bibitem{cao2018scale}
Xinkun Cao, Zhipeng Wang, Yanyun Zhao, and Fei Su,
\newblock ``Scale aggregation network for accurate and efficient crowd
  counting,''
\newblock in {\em ECCV}, 2018.

\bibitem{liu2019context}
Weizhe Liu, Mathieu Salzmann, and Pascal Fua,
\newblock ``Context-aware crowd counting,''
\newblock in {\em CVPR}, 2019.

\bibitem{wang2020distribution}
Boyu Wang, Huidong Liu, Dimitris Samaras, and Minh~Hoai Nguyen,
\newblock ``Distribution matching for crowd counting,''
\newblock {\em Advances in Neural Information Processing Systems}, vol. 33,
  2020.

\bibitem{lin2021direct}
Hui Lin, Xiaopeng Hong, Zhiheng Ma, Xing Wei, Yunfeng Qiu, Yaowei Wang, and
  Yihong Gong,
\newblock ``Direct measure matching for crowd counting,''
\newblock in {\em IJCAI}, 2021.

\bibitem{moranduzzo2013automatic}
Thomas Moranduzzo and Farid Melgani,
\newblock ``Automatic car counting method for unmanned aerial vehicle images,''
\newblock {\em TGRS}, 2013.

\bibitem{machefer2020mask}
M{\'e}lissande Machefer, Fran{\c{c}}ois Lemarchand, Virginie Bonnefond,
  Alasdair Hitchins, and Panagiotis Sidiropoulos,
\newblock ``Mask r-cnn refitting strategy for plant counting and sizing in uav
  imagery,''
\newblock {\em Remote Sensing}, 2020.

\bibitem{falk2019u}
Thorsten Falk, Dominic Mai, Robert Bensch, {\"O}zg{\"u}n {\c{C}}i{\c{c}}ek,
  Ahmed Abdulkadir, Yassine Marrakchi, Anton B{\"o}hm, Jan Deubner, Zoe
  J{\"a}ckel, Katharina Seiwald, et~al.,
\newblock ``U-net: deep learning for cell counting, detection, and
  morphometry,''
\newblock {\em Nature methods}, 2019.

\bibitem{xie2018microscopy}
Weidi Xie, J~Alison Noble, and Andrew Zisserman,
\newblock ``Microscopy cell counting and detection with fully convolutional
  regression networks,''
\newblock {\em Computer methods in biomechanics and biomedical engineering:
  Imaging \& Visualization}, 2018.

\bibitem{lu2018class}
Erika Lu, Weidi Xie, and Andrew Zisserman,
\newblock ``Class-agnostic counting,''
\newblock in {\em ACCV}, 2018.

\bibitem{michaelis2018one}
Claudio Michaelis, Ivan Ustyuzhaninov, Matthias Bethge, and Alexander~S Ecker,
\newblock ``One-shot instance segmentation,''
\newblock {\em arXiv preprint}, 2018.

\bibitem{hsieh2019one}
Ting-I Hsieh, Yi-Chen Lo, Hwann-Tzong Chen, and Tyng-Luh Liu,
\newblock ``One-shot object detection with co-attention and co-excitation,''
\newblock in {\em NIPS}, 2019.

\bibitem{li2020one}
Xiang Li, Lin Zhang, Yau~Pun Chen, Yu-Wing Tai, and Chi-Keung Tang,
\newblock ``One-shot object detection without fine-tuning,''
\newblock {\em arXiv preprint}, 2020.

\bibitem{lin2017focal}
Tsung-Yi Lin, Priya Goyal, Ross Girshick, Kaiming He, and Piotr Doll{\'a}r,
\newblock ``Focal loss for dense object detection,''
\newblock in {\em ICCV}, 2017.

\bibitem{ren2015faster}
Shaoqing Ren, Kaiming He, Ross Girshick, and Jian Sun,
\newblock ``Faster r-cnn: Towards real-time object detection with region
  proposal networks,''
\newblock {\em NIPS}, 2015.

\bibitem{he2017mask}
Kaiming He, Georgia Gkioxari, Piotr Doll{\'a}r, and Ross Girshick,
\newblock ``Mask r-cnn,''
\newblock in {\em ICCV}, 2017.

\bibitem{vaswani2017attention}
Ashish Vaswani, Noam Shazeer, Niki Parmar, Jakob Uszkoreit, Llion Jones,
  Aidan~N Gomez, {\L}ukasz Kaiser, and Illia Polosukhin,
\newblock ``Attention is all you need,''
\newblock in {\em NIPS}, 2017.

\bibitem{kingmaadam}
Diederik~P Kingma and Jimmy~Lei Ba,
\newblock ``Adam: Amethod for stochastic optimization,''
\newblock .

\bibitem{kang2019few}
Bingyi Kang, Zhuang Liu, Xin Wang, Fisher Yu, Jiashi Feng, and Trevor Darrell,
\newblock ``Few-shot object detection via feature reweighting,''
\newblock in {\em ICCV}, 2019.

\bibitem{fan2020few}
Qi~Fan, Wei Zhuo, Chi-Keung Tang, and Yu-Wing Tai,
\newblock ``Few-shot object detection with attention-rpn and multi-relation
  detector,''
\newblock in {\em CVPR}, 2020.

\bibitem{finn2017model}
Chelsea Finn, Pieter Abbeel, and Sergey Levine,
\newblock ``Model-agnostic meta-learning for fast adaptation of deep
  networks,''
\newblock in {\em ICML}, 2017.

\end{thebibliography}

\end{document}